\documentclass[letterpaper, 10 pt, conference]{ieeeconf}  

\usepackage{mathptmx} 
\usepackage{times} 
\usepackage{amsmath} 
\usepackage{amssymb}  
\usepackage{lipsum}

\usepackage{mathtools, nccmath}
\usepackage{cuted}

\usepackage{multirow}

\usepackage{paralist}
\usepackage{pifont} 
\usepackage{tabularx}
\usepackage{booktabs}
\usepackage{color}
\usepackage{colortbl}
\usepackage{color,soul}
\usepackage{float}
\definecolor{schwarz}{rgb}{0.0,0.0,0.0}
\definecolor{black}{rgb}{0.0,0.0,0.0}
\definecolor{darkdarkgray}{rgb}{0.6,0.6,0.6}
\definecolor{darkgray}{rgb}{0.8,0.8,0.8}
\definecolor{lightgray}{rgb}{0.95,0.95,0.95}
\definecolor{white}{rgb}{1.0,1.0,1.0}

\usepackage{stfloats}
\usepackage{blindtext}

\usepackage[export]{adjustbox}

\usepackage{epstopdf} 		 
\usepackage{epstopdf} 		 

\usepackage{ifluatex}
\ifluatex
\usepackage{pdftexcmds}
\makeatletter
\let\pdfstrcmp\pdf@strcmp
\let\pdffilemoddate\pdf@filemoddate
\makeatother
\fi
\usepackage{svg}

\usepackage{longtable}

\makeatletter
\DeclareFontFamily{U}{tipa}{}
\DeclareFontShape{U}{tipa}{m}{n}{<->tipa10}{}
\newcommand{\arc@char}{{\usefont{U}{tipa}{m}{n}\symbol{62}}}%

\newcommand{\arc}[1]{\mathpalette\arc@arc{#1}}

\newcommand{\arc@arc}[2]{%
	\sbox0{$\m@th#1#2$}%
	\vbox{
		\hbox{\resizebox{\wd0}{\height}{\arc@char}}
		\nointerlineskip
		\box0
	}%
}
\makeatother

\def\@fnsymbol#1{\ensuremath{\ifcase#1\or *\or \dagger\or \ddagger\or
		\mathsection\or \mathparagraph\or \|\or **\or \dagger\dagger
		\or \ddagger\ddagger \else\@ctrerr\fi}}

\newcommand{\ssymbol}[1]{{\@fnsymbol{#1}}}

\usepackage{graphicx}        
\usepackage{caption}

\usepackage{cite}

\usepackage{gensymb}
\def\°{^{\circ}}

\usepackage{algcompatible}
\usepackage{algorithm}
\usepackage{algpseudocode}
\algnewcommand{\LeftComment}[1]{\Statex \(\triangleright\) #1}

\makeatletter
\def\endthebibliography{%
	\def\@noitemerr{\@latex@warning{Empty `thebibliography' environment}}%
	\endlist
}
\makeatother

\title{\LARGE \bf
Hybrid Data-Driven and Analytical Model for Kinematic Control of a Surgical Robotic Tool}

\author{Francesco Cursi$^*$, \IEEEmembership{Student Member, IEEE}, Anh Nguyen \IEEEmembership{Member, IEEE}, Guang-Zhong Yang, \IEEEmembership{Fellow, IEEE}
	\thanks{		
		*Corresponding author
\newline
		The authors are with the Hamlyn Centre, Imperial College London, Exhibition Road, London, UK.
		\newline
		G.Z. Yang is also with the Institute of Medical Robotics, Shanghai Jiao Tong University, China
		\newline
		 Email:      
		[f.cursi17,g.z.yang]@imperial.ac.uk, gzyang@sjtu.edu.cn 
}
}
\IEEEoverridecommandlockouts

\begin{document}

	\maketitle
	\thispagestyle{empty}
	\pagestyle{empty}
		
	\begin{abstract}







Accurate kinematic models are essential for effective control of surgical robots. For tendon driven robots, which is common for minimally invasive surgery, intrinsic nonlinearities are important to consider. Traditional analytical methods allow to build the kinematic model of the system by making certain assumptions and simplifications on the nonlinearities. Machine learning techniques, instead, allow to recover a more complex model based on the acquired data. However, analytical models are more generalisable, but can be over-simplified; data-driven models, on the other hand, can cater for more complex models, but are less generalisable and the result is highly affected by the training dataset. In this paper, we present a novel approach to combining analytical and data-driven approaches to model the kinematics of nonlinear  tendon-driven surgical robots. Gaussian Process Regression (GPR) is used for learning the data-driven model and the proposed method is tested on both simulated data and real experimental data.
\end{abstract}

\section{Introduction}\label{sec:Introduction}



Robot kinematic modelling is is the pre-requisite of effective robot control. Having good kinematic models allows to properly control the robotic system, without requiring complex compensation strategies. The more accurate the robot model is, the more precise the control will be. Moreover, in cases where it is not possible to rely on external sensors to compensate for positioning errors (such as camera obstructions), accurate kinematic models are essential. 


There exist a large variety of robotic structures, such as rigid-link articulated robots, flexible-link robots, continuum robots, soft robots. Depending on the structure, different modelling techniques exist.
Articulated robots with rigid links are usually modelled by using Denavit-Hartenberg convention \cite{Siciliano2009RoboticsControl}.
Flexible link robots \cite{Cheong2004InverseApplications} are usually modelled by using Euler-Bernoulli beam theory and a set of generalized coordinates to describe the rigid and flexible motion.
The models of continuum robots and soft robots are generally derived by using constant-curvature, variable-curvature and Cosserat rod models \cite{Burgner-Kahrs2015}.\\

These models are often computed analytically, by means of mathematical formulations. However, analytical models are usually based on some assumptions and simplifications. Even though these simplifications allow to make the modelling easier and more understandable, they lead to modelling errors that need to be properly compensated by means of the control strategies \cite{Reinhart2017Hybrid}.
\begin{figure}
\vspace{5mm}
    \centering
    \includegraphics[width = \columnwidth]{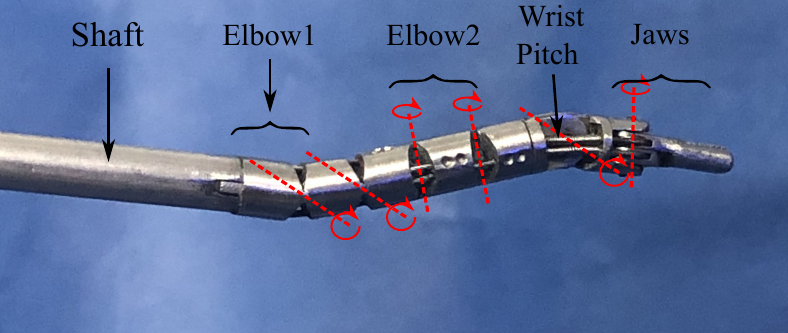}
    \caption{The Micro-IGES robotic surgical tool.}
    \label{fig:MicroIGES}
\end{figure}
Recently, there is a growing interest in the use of data-driven machine learning techniques for robot modelling. In the field of robotics, machine learning has been widely used to accurately approximate models of robots, without the need of analytical models, which may be hard to obtain due to the complexity of the system \cite{Nguyen-Tuong2011}. Despite proving very powerful, data-driven approaches depend on the algorithm chosen and, most of all, on the training data-set \cite{Nguyen-Tuong2011}. In order for the model to be accurate and generalizable, the training data must be gathered properly and should cover as much as possible the input and output spaces. Moreover, bad data points such as outliers should be rejected in order to avoid improper modelling \cite{Cursi2019ALearning}.\\

Robots for minimally invasive surgery have usually complex structures, being very articulated. Therefore, modelling their kinematics may be very challenging. Moreover, due to miniaturization requirements, flexibility, and sterilization, these robots are usually tendon-driven. Tendon transmission is a source of high nonlinearities due to hysteresis, tendon elongation and slack, friction. These nonlinearities are very hard to model, even if many researches have focused on building analytical models \cite{Do2015a,Ismail2009,Do2014c,Do2017,varghese2020nonlinearity,Tjahjowidodo2016,Palli2012,Do2013}. 
On the other hand, other works have focused on modelling robotic system by uing data-driven approaches.
Yu et al. \cite{YuProbabilisticControl} used Gaussian Mixture Model to build the kinematic model of a robotic catheter. A comparison of different approaches (Gaussian mixture models, k-nearest neighbour regression, and extreme machine learning) to model the inverse kinematics of a cable-robot was presented in \cite{Xu2017Data-drivenManipulators}. \\

However, combining analytical and data-driven models may leverage the advantages of both methods and thus improve the modelling. Thus far, little work has been focused on hybrid data-driven and analytical approaches.

Reinhart et al. \cite{Reinhart2017Hybrid} utilized three different data-driven approaches to be mixed with the analytical model of a soft robot based on the constant curvature assumption. The hybrid model is  built by using the data-driven approaches to learn the errors between the analytical model and the acquired data. 

Nguyen-Tuong et al. \cite{DuyNguyen-Tuong2010UsingDynamics} incorporated the known dynamic model of a robot into the prior of a Gaussian Process, either by using it in the process mean function or in the kernel function. In both works, the hybridization yields to better modelling results, with improvements also in the generalization of the model.\\
 
To the best of the authors knowledge, these hybridization methods have little been applied in the field of minimally invasive-surgery, where high accuracy is required. For instance, \cite{Porto2019PositionAnalysis} exploited the knowledge of the kinematic model of a continuum robot to improve the learning of the inverse kinematics, compensating for hysteresis. This, however, required learning different forward and inverse models and the robot was supposed to bend following the constant curvature assumption.

In this work we present a novel approach for combining analytical and data-driven approaches for modelling the forward kinematics of robots, with particular emphasis to the Micro-IGES \cite{Shang2017} (Figure \ref{fig:MicroIGES}), a tendon-driven robotic surgical tool. The method utilizes Gaussian Process Regression (GPR) for the computation of the data-driven model. This regression method has been chosen thanks to its ability to provide a confidence interval, indicating the uncertainty in the model.  The contribution of the paper is therefore two-fold:
 \begin{itemize}
      \item compare the results for kinematic modelling by using different approaches based on GPR;
     \item present a novel method for kinematic modelling based on mixing data-driven and analytical methods.\\
 \end{itemize}

The paper is therefore structured as follows.
Section \ref{sec:ModelLearning} presents the robot under exam and describes the computation of the analytical, data-driven, and hybrid models, along with a brief introduction of Gaussian Process Regression. Section \ref{sec:Results} presents the results for the forward kinematic modelling of the system. Different data-driven approaches are used for the modelling and their results are compared to those of the proposed hybrid approach. The methods are tesetd both on simulated and real data.  Finally, conclusions are drawn in Section \ref{sec:Conclusions}.
	\section{Robot Forward Kinematic Models}\label{sec:ModelLearning}
The kinematic model of a robot allows to estimate the end-effector Cartesian pose (position and orientation), given some input joint values. In tendon driven system, however, the tendon transmission leads to nonlinearities in the motor to joint mapping, which are difficult to estimate analytically, thus requiring data-driven approaches to estimate the robot pose. In this Section a brief overview of the robot under consideration is presented. Then, the computation of the data-driven and analytical kinematic models are described. Finally, the approach to combine the two models to obtain a hybrid one is presented.

\subsection{Gaussian Process Regression}
Gaussian Process Regression is a nonparametric regression method that allows to approximate any nonlinear function $y_i \sim f(\mathbf{x_i})+v_i$, where $y_i$ with $i = 1...N$ is the vector of measured outputs, $\mathbf{x_i} \in \mathbb{R}^{n_{in}}$ is the input vector, and $v_i$ is Gaussian noise with $0$ mean and variance equal to $\sigma_y^2$ \cite{Murphy2012a,Rasmussen2006}.

A Gaussian Process $f(x) \sim GP(m(x),k(x,x'))$ is completely defined by its mean function $ m(x)$ and covariance function $ k(x,x')$, and it defines the $prior$ distribution of the data. Given the knowledge of the prior on the training set defined by $\mathbf{X} \in \mathbb{R}^{n_{in}\times N}$ and $\mathbf{y} \in \mathbb{R}^N$, it is possible to estimate the $posterior$ distribution on a test input set $\mathbf{X_*}$ which results to be
\begin{equation}\label{eq:GP}
\begin{aligned}
   & p(f_*|\mathbf{X,X_*,y}) \sim \mathcal{N}(\mathbf{\mu}, \Sigma)\ , \\
   &\mathbf{\mu = m(X_*)+K(X_*,X)K_y^{-1}(y-m(X))}\ ,\\
   & \Sigma = \mathbf{K(X_*,X_*)-K(X_*, X)K_y^{-1}K(X, X_*)}\ ,\\
   &\mathbf{P(\theta) = m(\theta)+K_{*}^{T}(\theta,\tilde{X})K_y^{-1}(\tilde{P}-m(\tilde{X}))}\\
   & \sigma^2 = \mathbf{K(\theta,\theta)-K_{*}^{T}(\theta,\tilde{X})K_y^{-1}K_{*}(\theta,\tilde{X})}\\
   & u = \frac{1}{2}\sigma^{2^T}\sigma^{2}\\
   &\mathbf{\dot{q} = J^\ssymbol{2}v+(I-J^\ssymbol{2}J)\dot{q}_0}\\
   & \mathbf{J = \frac{\partial P}{\partial \theta}} \\
   & \mathbf{\dot{q}_0 = -\frac{\partial \sigma^2}{\partial \theta}}\frac{\sigma{^2}}{dt}
\end{aligned}
\end{equation}
where $\mathbf{K}$ is the covariance matrix between the input values and $\mathbf{K_y = K(X, X) + \sigma_y^2I }$.

The prior knowledge of the model can therefore be incorporated in the regression by defining the mean function $\mathbf{m(x)}$ of the Gaussian process \cite{Rasmussen2006}.

\subsection{The Micro-IGES Robotic Surgical Tool}
The Micro-IGES \cite{Shang2017,Leibrandt2017} (Fig. \ref{fig:MicroIGES}) is a surgical robotic tool, composed of a rigid shaft (27 cm) and a flexible section (39mm at zero configuration). The shaft is responsible for the roll and translation DOFs. The articulated end, instead, consists of 2 elbows for pitch and yaw, with each elbow made of a pair of coupled joints, a 1DOF revolute joint for the wrist pitch, and the jaws. The jaws provide two more DOFs: one for the wrist yaw and one for the gripper's opening/closing. Each of the joints of the articulated part is driven by an antagonistic pair of tendons, with each pair being connected to the corresponding driving capstan at the proximal drive unit. The coupling of the two pairs of joints of the elbows occurs at the driving unit: the two capstans that drive the two serial joints for each DOF of the elbow (pitch and yaw) are coupled by a series of gears with 1:2 ratio.\\

\subsection{Analytical Kinematic Model}
In many applications models of the kinematics of the robots have been developed analytically. In order to control a robot, desired motor input values must be provided. These motor values are converted, by means of the motor transmission, to joint values, which, in turn, depending on the kinematic model of the robot, are then mapped onto the end-effector Cartesian pose. The analytical approach consists in finding mathematical relationships between the  joint values and the the Cartesian pose, and between the motor values and the joint values.

The joint to Cartesian mapping depends on the geometry of the robot. For articulated robots, Denavit-Haretnber convention can be used \cite{Siciliano2008}. In continuum robots, instead, constant curvature or variable curvature models have been developed \cite{Burgner-Kahrs2015,Camarillo2008,Bajo2012}.
The motor to joint mapping, instead, depends on the type of motor transmission. In tendon-driven systems analytical models of hystersis, friction, tendon elongation need to be formulated \cite{Do2015a,Ismail2009,Do2014c,Do2017, varghese2020nonlinearity, Tjahjowidodo2016,Palli2012,Do2013}.

Being the Micro-IGES robot an articulated robot, the joint to tip-pose mapping is computed by means of Denavit-Hartenberg convention. For the motor to joint mapping, an hysteresis model is included as described in \cite{Leibrandt2017}.
The analytical model is then described as 
\begin{equation}
\begin{aligned}
    & \mathbf{P = P_a}(\Theta_a)\ ,\\
    &\Theta_a = \begin{bmatrix} \theta & \dot{\theta} & \dot{\theta}_{old} \end{bmatrix}^T\ .
\end{aligned}
\end{equation}
where $\mathbf{P} \in \mathbb{R}^3$ is the end-effector Cartesian position, $\mathbf{\theta, \dot{\theta}} \in  \mathbb{R}^{n_m}$ describe the current motor state and $\mathbf{\dot{\theta}_{old}}$ is the motor velocity in the previous state. The addition of the actual and past motor velocities are needed to compensate for the hysteretical behaviour.

\subsection{Data-Driven Kinematic Model}
Despite being very generalizable, analytical models are usually based on some approximations and assumptions. The errors in the modelling may lead to wrong or poor control. In tendon driven systems, especially, the nonlinearities in the motor to joint mapping are very difficult to model analytically, and the analytical approximations may not be satisfying enough.

Data-driven approaches, on the other hand, allow to build models of the system based on the data acquired.
For the computation of the data-driven model, Gaussian process regression is used thanks to its ability of providing a confidence interval of the model.\\

In order to compensate for the nonlinearities in the Micro-IGES system, the data-driven model is computed as
\begin{equation}
\begin{aligned}
    & \mathbf{P = P_d}(\Theta_d)\ ,\\
    &\Theta_d = \begin{bmatrix} \theta & \dot{\theta} & \ddot{\theta}& \theta_{old} & \dot{\theta}_{old} \end{bmatrix}^T\ .
\end{aligned}
\end{equation}
The input vector includes the motor velocities and accelerations in order to have better compensation of hysteretical effects and friction.
In order to compute $\mathbf{P_d}$, three different independent Gaussian Processes need to be used. Therefore, $\mathbf{P_d} = \begin{bmatrix} \mu_x & \mu_y & \mu_z \end{bmatrix}^T $ with each $\mathbf{\mu}$ corresponding to each posterior mean value obtained from the Gaussian Processes. Each predicted position component is also associated with a variance, defining $\mathbf{\sigma_d^2} = \begin{bmatrix} \sigma_x^2 & \sigma_y^2 & \sigma_z^2 \end{bmatrix}^T$. LimboGP library \cite{Cully2016Limbo:Optimization} for C++ has been used for the Gaussian Process Regression.

\subsection{Hybrid Kinematic Model}


Both analytical and data-driven methods have their advantages and disadvantages. Analytic models are typically more generalizable and can be applied to different scenarios. Nevertheless, they are often based on simplifications. Data-driven models, instead, allow for more complex modelling, but they rely on the acquired data. This makes the generalization more difficult, due to poor dataset exploration, and may lead to wrong models if data contains outliers \cite{Cursi2019ALearning}.
Therefore, to improve the accuracy of the model, a combination of the two may be necessary.\\
Let $\Theta = \begin{bmatrix} \theta & \dot{\theta} & \ddot{\theta} & \theta_{old} & \dot{\theta}_{old} \end{bmatrix}^T \in \mathbb{R}^{5n_j}$ be a single input vector, $\mathbf{P_d}(\Theta)$ the data-driven model associated with $\mathbf{\sigma_d^2}(\Theta)$, and $\mathbf{P_a}(\Theta)$ the analytical model. The hybrid model is then computed as (\ref{eq:KineModel})
\begin{equation}\label{eq:KineModel}
\mathbf{P}(\Theta)  =  \mathbf{(I-W) P_d}(\Theta)+\mathbf{ W P_a}(\Theta)\ ,
\end{equation}
where $\mathbf{W} \in \mathbb{R}^{3\times 3}$ is a weighting diagonal matrix, with $W_{i} \in [0\ 1]$. For the computation of the analytical model the input value can be computed as $\Theta = \begin{bmatrix} \theta & \dot{\theta} & 0 & 0 & \dot{\theta}_{old} \end{bmatrix}^T $.

Each component $i = 1,2,3$ of $\mathbf{W}$ is computed as (\ref{eq:weight})
\begin{equation}\label{eq:weight}
    W_i = e^{-k_i \frac{(P_{d,i}-P_{a,i})^2}{\sigma_{d,i}^2}}\ .
\end{equation}
Equation (\ref{eq:weight}) shows that in regions where the data-driven model is more uncertain (high variance) the weight tends to 1, thus favouring the analytical model. On the contrary, if the uncertainty of the data-driven model is low, the data-driven model is preferred. Moreover, if the error between the two models is high, then the data-driven model is preferred.

The value $k_i$ is defined as $\frac{|P_{d,i}-P_{a,i}|}{t}$, where $t$ is a desired threshold. This threshold can have different values for each Cartesian component. This regularization is needed because, if the data-driven model is very uncertain, the analytical approach may be favoured, even if the error is high. Therefore, large values of $k_i$ will tend to give more importance to the data-driven model.  




	\section{Results}\label{sec:Results}
\begin{figure}[h]
\vspace{5mm}
    \centering
    \includegraphics{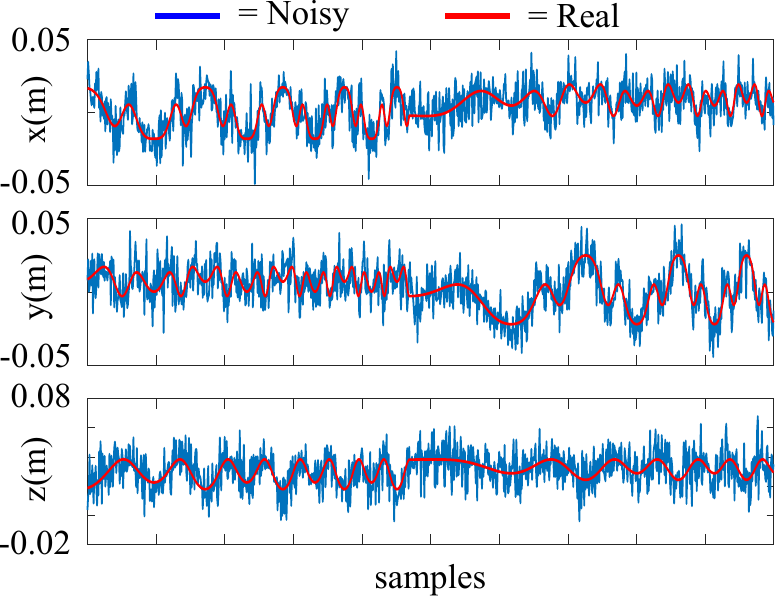}
    \caption{Window showing the noisy data used for the learning process in simulation. The red line is the actual data; the blue line is the data with added noise used for the learning.}
    \label{fig:Learndata}
\end{figure}


\begin{figure*}
\vspace{5mm}
    \centering
    \includegraphics{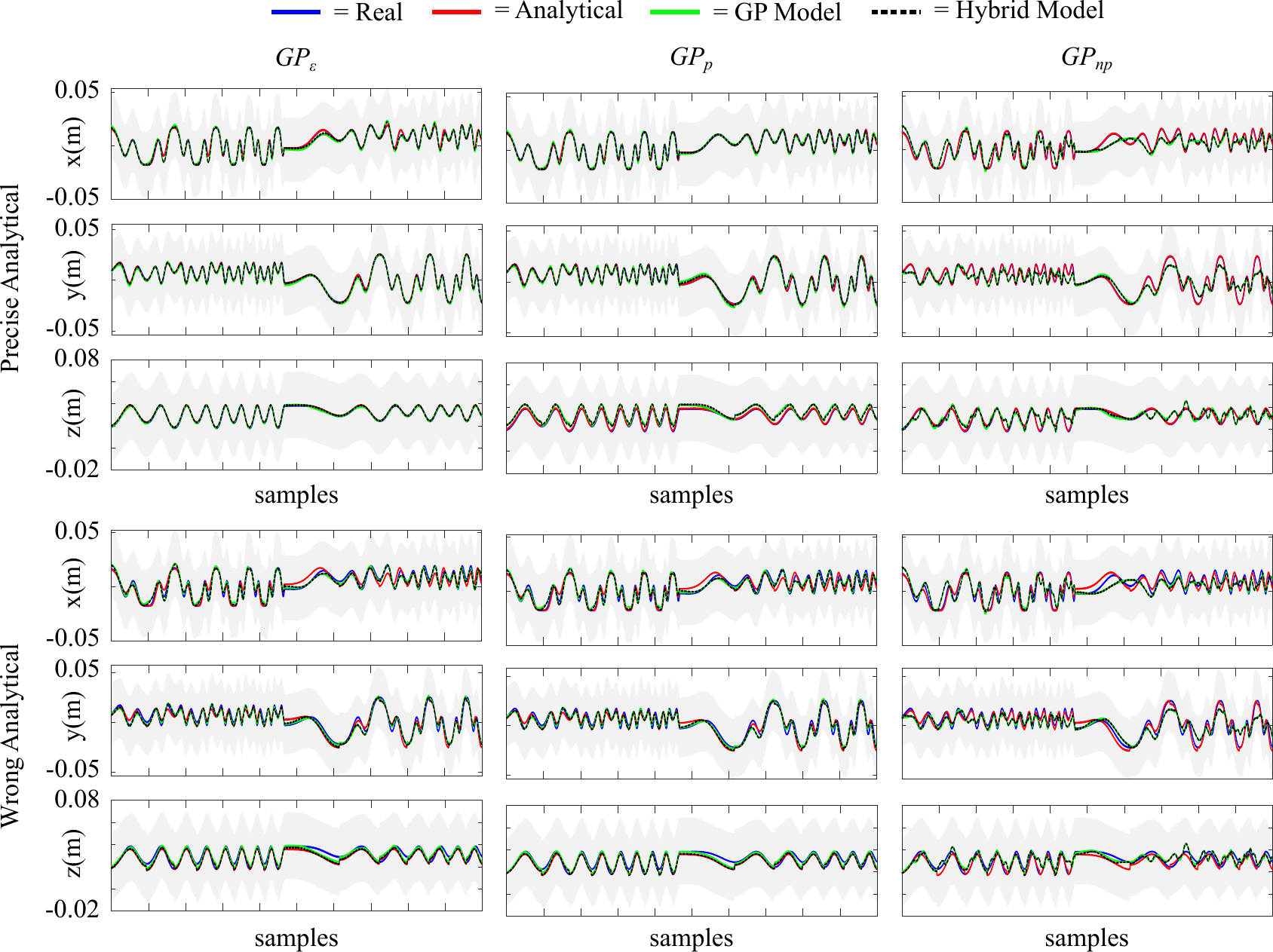}
    \caption{Model comparison on a subset of the simulated learning data with both a precise and a wrong analytical model. For each case, the three different GPR and hybridization approaches are shown. The shaded regions indicate a 95\% confidence interval of the GP models.}
    \label{fig:ResFKine_Sim}
\end{figure*}

In this Section the results for modelling the kinematics of the Micro-IGES robotic surgical tool are presented. Different approaches are compared for the computation of the data-driven models. Moreover, two different analytical models (a wrong model and a more precise one) are used to test the results from the hybrid models.

\subsection{Simulated Forward Kinematic Model}\label{subsec:FwdKineRes}
The tendon-transmission in the Micro-IGES robot causes many nonlinearities due to friction, hysteresis, tendon elongation, etc. To validate the proposed method a simulated environment in VREP \cite{VREP} is used. The robot is supposed to exhibit linear backlash at the motor side. Data for the learning process are acquired by commanding excitation trajectories to the robot and retrieving the tip position from the simulator. Each motor $i = 1...n_m$ is excited as (\ref{eq:excite})
\begin{equation}\label{eq:excite}
\begin{aligned}
&\theta_i(t) = h_i( a_i \sin(\omega t + \psi_i)+b_i \cos(\omega t +\phi_i)+\theta_{0,i})\\
& \text{with}\ \omega = \omega_{min}+\frac{\omega_{max}-\omega_{min}}{T} t\ ,
\end{aligned}
\end{equation}
where $\omega_{min} = 0.1 Hz, \omega_{max} = 1 Hz$, $T = \frac{\pi}{\omega_{min}}$. The parameters $a_i,b_i,\psi_i,\phi_i$,$\theta_{0,i}$ are computed by solving an optimization problem, maximizing the amplitude of the wave while satisfying position, velocity, and acceleration limit for each motor during the whole motion. For each motor $h_i$ is either 0 or 1, depending if the desired motor is moving or not. In total $2^{n_m}$ combinations of motions are performed.

Some noise is then added to each Cartesian component. The noise is Gaussian with zero mean and standard deviation of $0.01 m$. Figure \ref{fig:Learndata} shows a window of the data used for the learning. In total 19499 samples were collected. For the learning, however, 1000 random samples are used due to limitations due to the GPR.

In order to learn the kinematic model of the robot three different approaches are considered for the data-driven models. Consequently, three different hybrid models are also computed:
\begin{itemize}
    \item \textbf{Error Learning ($GP_{\epsilon}$)}: the error between the analytical model and the data is learned by using GPR.
    
    In this case the data-driven model can be written as $\mathbf{P_d = P_a+\epsilon_{GP}}$, where $\mathbf{\epsilon_{GP}}$ is the error between the data and the analytical model learned through GPR. The hybrid model, according to (\ref{eq:KineModel}) results to be $\mathbf{P = P_a+(I-W)\epsilon_{GP}}$. This means that the error compensation is activated or deactivated based on the level of uncertainty. 
    
    \item \textbf{GP with Prior ($GP_{p}$)}: the analytical model is used as mean function for the prior distribution.
    Form (\ref{eq:GP}), the data-driven model is computed as $\mathbf{P_d = P_a+W_{GP}(y-\tilde{P}_a)}$ \cite{DuyNguyen-Tuong2010UsingDynamics}, where $\mathbf{W_{GP}=K(X_*,X)K_y^{-1}}$ and $\mathbf{\tilde{P}_a}$ is the value of the analytical model on the training set. The resulting hybrid model is then $\mathbf{P = P_a+(I-W)W_{GP}(y-\tilde{P}_a)}$. In this case, the contribution of the prior distribution is affected not only by the values in the input space (by means of the covariance matrices), but also by the output values.
    
    \item \textbf{GP without Prior ($GP_{np}$)}: the input output mapping is computed without any prior knowledge. The hybrid model is computed as $\mathbf{P = (I-W)P_d+WP_a}$.

\end{itemize}

\begin{table}[t]
\vspace{5mm}
	\caption{RMSE (in m) between the computed models and the ground truth of the simulated model on the learning dataset.}
	\centering
	\begin{tabular}{|c|c|c|c|c|c|c|}
			\hline
		\multicolumn{7}{|c|}{\textbf{Wrong Analytical}}\\
		\hline 
		\multicolumn{7}{|c|}{\textbf{RMSE wrt ground truth}}\\
		\hline 
		& $GP_{\epsilon}$ & $GP_{p}$ & $GP_{np}$ & $Hyb_{\epsilon}$ & $Hyb_{p}$ & $Hyb_{np}$\\
		\hline
		{x}
		 & \textbf{0.0022} & 0.0024 & 0.0041 & 0.0025  & 0.0026 & 0.0041\\
		 
		  {y}
		 & \textbf{0.0024} & 0.0027 &0.0045 & 0.0026 & 0.0029 & 0.0044\\
		 
		  {z}
		 & \textbf{0.0016} & 0.0018 & 0.0037 & 0.0022 & 0.0023 & 0.0035\\
		\hline
		\hline
		
				\multicolumn{7}{|c|}{\textbf{RMSE wrt analytical model}}\\
		\hline 
		& $GP_{\epsilon}$ & $GP_{p}$ & $GP_{np}$ & $Hyb_{\epsilon}$ & $Hyb_{p}$ & $Hyb_{np}$\\
		\hline
		{x}
		 & 0.0039 & 0.0038 & 0.0055 & 0.0032  & \textbf{0.0032} & 0.0052\\
		 
		  {y}
		 & 0.0037 & 0.0035 &0.0057 & 0.0031 & \textbf{0.0029} & 0.0054\\
		 
		  {z}
		 & 0.0023 & 0.0021 & 0.0043 & 0.0008 & \textbf{0.0007} & 0.0038\\
		\hline
		\hline
		\hline
		
		\multicolumn{7}{|c|}{\textbf{Correct Analytical}}\\
		\hline 
		\multicolumn{7}{|c|}{\textbf{RMSE wrt ground truth}}\\
		\hline 
		& $GP_{\epsilon}$ & $GP_{p}$ & $GP_{np}$ & $Hyb_{\epsilon}$ & $Hyb_{p}$ & $Hyb_{np}$\\
		\hline
		{x}
		 & 0.0021 & 0.0021 & 0.0041 & \textbf{0.0014}  & 0.0014 & 0.0037\\
		 
		  {y}
		 & 0.0019 & 0.0027 &0.0045 & \textbf{0.0010} & 0.0017 & 0.0041\\
		 
		  {z}
		 & 0.0008 & 0.0018 & 0.0037 & \textbf{0.0007} & 0.0012 & 0.0032\\
		\hline
		\hline
		
				\multicolumn{7}{|c|}{\textbf{RMSE wrt analytical model}}\\
		\hline 
		& $GP_{\epsilon}$ & $GP_{p}$ & $GP_{np}$ & $Hyb_{\epsilon}$ & $Hyb_{p}$ & $Hyb_{np}$\\
		\hline
		{x}
		 & 0.0020 & 0.0018 & 0.0041 & 0.0012  & \textbf{0.0010} & 0.0037\\
		 
		  {y}
		 & 0.0018 & 0.0022 &0.0045 & \textbf{0.0007} & 0.0012 & 0.0042\\
		 
		  {z}
		 & 0.0007 & 0.0016 & 0.0037 & \textbf{0.00005} & 0.0009 & 0.0032\\
		\hline
		
	\end{tabular}
	
	\label{tab:results_sim}
\end{table}

\begin{table}[t]
	\caption{Maximum absolute error (in m) between the computed models and the simulated model on the learning dataset.}
	\centering
	\begin{tabular}{|c|c|c|c|c|c|c|}
			\hline
		\multicolumn{7}{|c|}{\textbf{Wrong Analytical}}\\
		\hline 
		& $GP_{\epsilon}$ & $GP_{p}$ & $GP_{np}$ & $Hyb_{\epsilon}$ & $Hyb_{p}$ & $Hyb_{np}$\\
		\hline
		{x}
		 & 0.0085 & 0.0092 & 0.0201 & \textbf{0.0073}  & 0.0089 & 0.0201\\
		 
		  {y}
		 & 0.0096 & 0.0098 &0.0236 & \textbf{0.0083} & 0.0087 & 0.0236\\
		 
		  {z}
		 & 0.0056 & 0.0060 & 0.0155 & \textbf{0.0056} & 0.0069 & 0.0155\\
		\hline
		\hline
		
		\multicolumn{7}{|c|}{\textbf{Correct Analytical}}\\
		\hline 
		& $GP_{\epsilon}$ & $GP_{p}$ & $GP_{np}$ & $Hyb_{\epsilon}$ & $Hyb_{p}$ & $Hyb_{np}$\\
		\hline
		{x}
		 & 0.0072 & 0.0011 & 0.0201 & \textbf{0.0066}  & 0.0011 & 0.0201\\
		 
		  {y}
		 & 0.0062 & 0.0096 &0.0236 & \textbf{0.0052} & 0.0096 & 0.0236\\
		 
		  {z}
		 & 0.0028 & 0.0054 & 0.0155 & \textbf{0.0026} & 0.0047 & 0.0155\\
		\hline
	
	\end{tabular}
	
	\label{tab:results_Max_error_sim}
\end{table}

The results of the proposed method are then tested on two different cases: with a wrong analytical model, and with a more precise analytical model. Figure \ref{fig:ResFKine_Sim} and Table \ref{tab:results_sim} show the results for the different approaches on the learning dataset.\\

In the case of wrong analytical model, no backlash compensation is used and wrong links' lengths are assumed.
In the second case, instead, correct measures and the same backlash compensation used for gathering the data are considered. In both cases, the threshold value in the weighting function (\ref{eq:weight}) has been set to $0.5mm$ for each Cartesian component.

Comparing the three data-driven methods, results show that learning the error ($GP_\epsilon$) yields to better results, with smaller RMSE on both datasets. The pure data-driven ($GP_{np}$), instead, is the one that always performs the worst. 

When hybridization is used, adding the analytical model as in (\ref{eq:KineModel}), results vary depending on the accuracy of the model provided. If a wrong analytical model is used, the RMSE errors between the computed models and the real one are slightly higher than in the case without hybridization. As a matter of fact, the hybridization with the chosen parameters (weigthing function, threshold, process confidence interval) makes the model tend more toward the provided analytical model. Also in this case, though, the hybrid model $Hyb_\epsilon$ built from $GP_{\epsilon}$ is the one that performs the best, whereas the one from $GP_{np}$ the worst. However, as shown in Table \ref{tab:results_Max_error_sim} the maximum absolute error is reduced (or at least not increased) for all the models by adding the hybridization.

If, on the other hand, a more accurate model is used, the hybridization leads to much better results. The RMSE are much smaller than in the case without hybridization. As expected, also the maximum absolute errors decrease. \\

In order to further validate the proposed method, the robot was required to perform an additional testing motion. The motion is described by:
\begin{equation}
\begin{aligned}
&\theta_i = 
\begin{cases}
\theta_{M,i} s(t)\ ,& t\in [0,T] \\
(\theta_{m,i}-\theta_{M,i}) s(t-T)+\theta_{M,i}\ ,& t\in (T,2T] \\
-\theta_{m,i} s(t-2T)+\theta_{m,i}\ ,& t\in (2T,3T] 
\end{cases}\\
&s(t) = 6\Big(\frac{t}{T}\Big)^5-15\Big(\frac{t}{T}\Big)^4+10\Big(\frac{t}{T}\Big)^3\ \in [0,1],
\end{aligned}
\end{equation} 
with $T = 5 s$ and $\theta_{m,i},\theta_{M,i}$ being the maximum and minimum motor position values.
The end-effector tip position was collected during the motion. In this case, the hybridization was performed with the more precise analytical model.
Results in Table \ref{tab:results_test_motion} show again that learning the error has better results than the other data-driven approaches. Moreover, the hybridization leads to improved results for all cases.

\begin{table}[h]
\vspace{2mm}
	\caption{RMSE and maximum absolute error (both in m) between the computed models and the ground truth of the simulated model on the testing motion.}
	\centering
	\begin{tabular}{|c|c|c|c|c|c|c|}
			\hline
		\multicolumn{7}{|c|}{\textbf{RMSE wrt ground truth}}\\
		\hline 
		& $GP_{\epsilon}$ & $GP_{p}$ & $GP_{np}$ & $Hyb_{\epsilon}$ & $Hyb_{p}$ & $Hyb_{np}$\\
		\hline
		{x}
		 & 0.0025 & 0.0024 & 0.0040 & \textbf{0.0018}  & 0.0015 & 0.0034\\
		 
		  {y}
		 & 0.0018 & 0.0032 &0.0056 & \textbf{0.0012} & 0.0025 & 0.0053\\
		 
		  {z}
		 & 0.0005 & 0.0011 & 0.0036 & \textbf{0.0005} & 0.0008 & 0.0029\\
		\hline
		\hline
		
		\multicolumn{7}{|c|}{\textbf{Maximum Absolute Error wrt gound truth}}\\
		\hline 
		& $GP_{\epsilon}$ & $GP_{p}$ & $GP_{np}$ & $Hyb_{\epsilon}$ & $Hyb_{p}$ & $Hyb_{np}$\\
		\hline
		{x}
		 & 0.0071 & 0.0051 & 0.0090 & \textbf{0.0056}  & 0.0037 & 0.0088\\
		 
		  {y}
		 & 0.0047 & 0.0085 &0.0124 & \textbf{0.0041} & 0.0084 & 0.0124\\
		 
		  {z}
		 & 0.0012 & 0.0022 & 0.0076 & \textbf{0.0012} & 0.0012 & 0.0076\\
		\hline
	
	\end{tabular}
	
	\label{tab:results_test_motion}
\end{table}

\subsection{Real Data Acquisition}\label{subsec:DataAcq}
In order to build the kinematic model of the real robot, the tool tip position needs to be acquired. For this purpose, a Intel Realsense stereo camera has been used. The exciting motor trajectories commanded for the data acquisition are similar to those in the simulated experiments in (\ref{eq:excite}), but with maximum frequency of $0.5\ Hz$. Also, only 4 degrees of freedom are considered (Roll, Elbow1, Elbow2, Wrist Pitch). For learning the data-driven model, 1000 samples are used, randomly taken from the whole collected dataset.

\begin{figure}[h]
    \centering
    \includegraphics[width = \columnwidth]{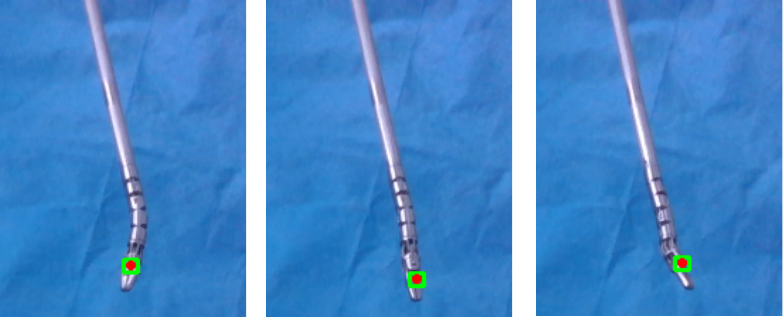}
    \caption{Snapshot of the motion for the data collection tracking the tip position.}
    \label{fig:Vision}
\end{figure}

We employ the CSRT Tracker\cite{Lukezic2016DiscriminativeReliability} to track the tip position of the robot from 2D images in real-time. This 2D position is then projected into 3D using the associated depth images and the focal length of the depth camera as follows:
\begin{equation}
    \begin{array}{l}
u = (x - c_x)*{D_{x,y}}/f_x\\
v = (y - c_y)*{D_{x,y}}/f_y\\
w = {D_{x,y}}
\end{array}
\end{equation}
where $x,y$ is the tip position in 2D image; $D_{x,y}$ is the depth value from associated depth image; $f_x$, $f_y$, and $c_x$, $c_y$ are the intrinsic parameters of the camera. Figure \ref{fig:Vision} shows a snapshot of the motion during data collection. 

The collected tip positions, however, are expressed in the camera frame. Since all the measurements need to be expressed in the robot base frame, camera calibration is used to map the collected tip position from the camera frame to the robot frame. 

\subsection{Real Robot Kinematic Model}\label{subsec:RealExp}
\begin{figure*}
\vspace{5mm}
    \centering
    \includegraphics[width = \textwidth]{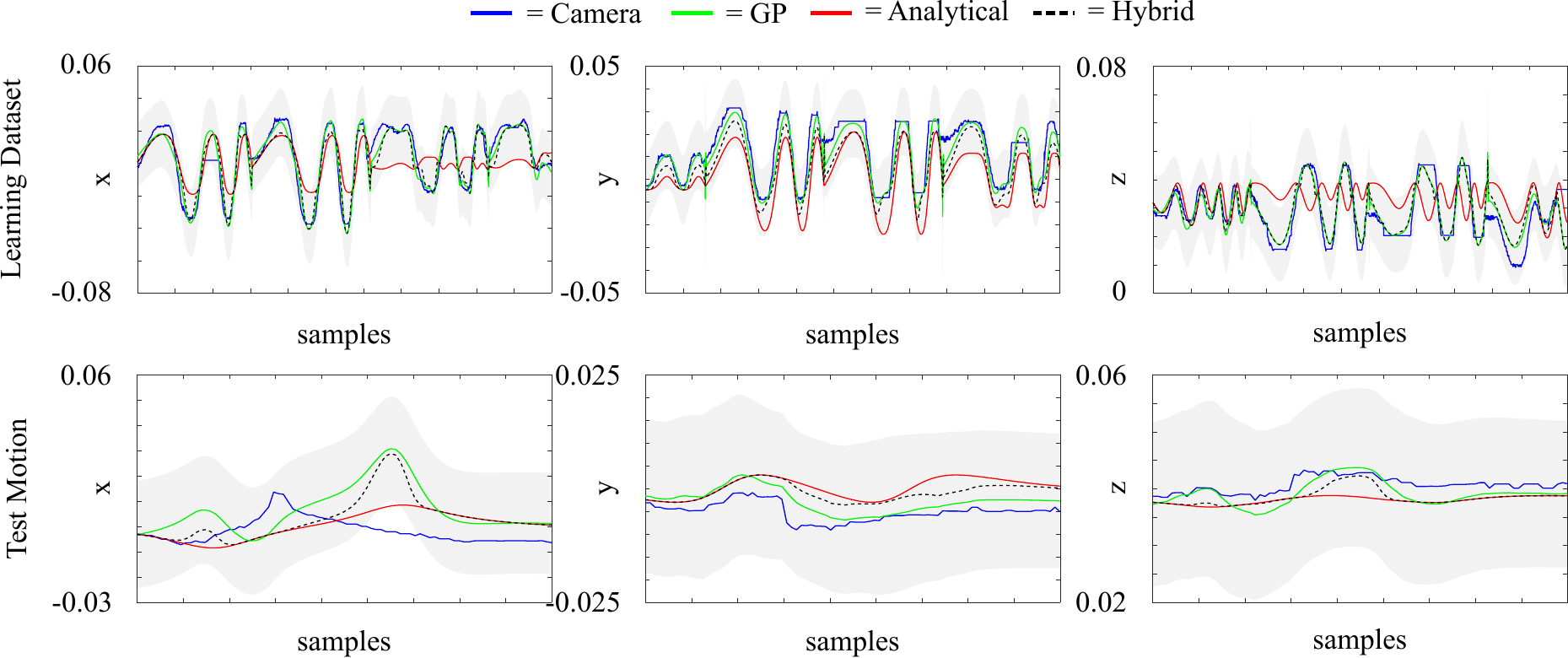}
    \caption{Comparison of the different models on a subset of the dataset from the real experiment, both for the learning (upper row) and the test motion (bottom row). All units are in m. The shaded regions indicate a 95\% confidence interval of the GP models.}
    \label{fig:ResReal}
\end{figure*}
Due to the better results obtained from learning the error ($GP_\epsilon$) between the analytical and the collected data, this approach is employed for retrieving the data-driven model of the real robot. For the hybridization, the precise analytical model (with backlash compensation and correct links' lengths) is employed and the thresholds $t$ in (\ref{eq:weight}) are chosen as $\begin{bmatrix}0.10& 0.01 &0.005 \end{bmatrix} m$, respectively for $x,y,z$. The upper row of Figure \ref{fig:ResReal} shows the results comparing the different models (analytical, GP, hybrid) on a subset of the collected learning dataset. 

As expected, due to unmodelled nonlinearities, the analytical model does not always behave properly. When some complex motion is commanded, for instance when more joints are moving together, the analytical model differs largely from the collected data. Under some simpler motions, instead, the analytical model results satisfactory.
The data-driven model allows to explain pretty well the collected data, with a RMSE for each component of $\begin{bmatrix}0.0068& 0.0059& 0.0047\end{bmatrix}m$ and a maximum absolute error of $\begin{bmatrix}0.0039& 0.0075& 0.0052\end{bmatrix}m$.  

Nevertheless, relying solely on the data-driven model may lead to wrong behaviours, due to errors in the camera calibration or noise in the collected data. When hybridization is added, the RMSE of the hybrid model with respect to the collected data increases, resulting to be $\begin{bmatrix}0.0089& 0.0060& 0.0048\end{bmatrix}m$. However, the maximum absolute error is kept invariant. This is because in regions where the analytic model is good, it is preferred or, at least, it has influence on the hybrid model. Otherwise, the hybrid model is always closer to the data-driven model, due to the inability of the analytical to model the system appropriately. \\

To further validate the models, a test motion is also performed. The motor values to command to the motors are computed from the analytical model imposing a Cartesian tip trajectory described by the lemniscate of Bernoulli path as
\begin{equation}
\begin{aligned}
& x(t) = 6\cdot 10^{-3} \frac{\sqrt{2}}{\sin^{2}(s(t))+1}\cos(s(t))+x_i \ ,\\
& y(t) =  6\cdot 10^{-3} \frac{\sqrt{2}}{\sin^{2}(s(t))+1}\frac{\sin(2s(t))}{2}+y_i \ ,\\
& roll(t) = (\alpha_f-\alpha_i)\sin(s_{roll}(t)\pi)+\alpha_i, \\
& s(t) = \Big[ 24\Big(\frac{t}{T}\Big)^{5}-60\Big(\frac{t}{T}\Big)^4+40\Big(\frac{t}{T}\Big)^3 +1\Big]\frac{\pi}{2}\ \in [\frac{\pi}{2},\frac{5\pi}{2}] \\
& s_{roll}(t) = 6\Big(\frac{t}{T}\Big)^{5}-15\Big(\frac{t}{T}\Big)^4+10\Big(\frac{t}{T}\Big)^3\ \in [0,1]\ .
\end{aligned}
\end{equation}
The initial and final configurations are set to $\mathbf{P_i} = \begin{bmatrix}0& 0& 0.038\end{bmatrix}^Tm$, $\alpha_i = 0$, $\alpha_f = 100\°$, and the execution time to $T = 6s$.
For the hybridization, the same thresholds as in the learning experiments are used. The lower row of Figure \ref{fig:ResReal} shows the results on a subset of the data.
Also in this case, the analytical model doesn't manage to provide accurate results, especially for the $x$ direction. The RMSE and maximum absolute errors result to be $\begin{bmatrix}0.0064& 0.0051& 0.0024\end{bmatrix}m$ and $\begin{bmatrix}0.016& 0.010& 0.0046\end{bmatrix}m$. However, also the GP model doesn't appear very accurate with the RMSE and maximum error being $\begin{bmatrix}0.011& 0.002& 0.0017\end{bmatrix}m$ and $\begin{bmatrix}0.033& 0.0067& 0.0039\end{bmatrix}m$. The larger error values in the learned model indicate the low ability of the data-driven approach to generalize, with the provided learning dataset.
When the hybridization is included, the resulting model is a bit smoother than the data-driven model, yielding to RMSE and maximum error values of $\begin{bmatrix}0.009& 0.0041& 0.002\end{bmatrix}m$ and $\begin{bmatrix}0.031& 0.0096& 0.0046\end{bmatrix}m$. This shows that where the analytic model behaves well, the hybridization improves the performances of the data-driven approach, as in the $x$ direction. Conversely, when it behaves poorly, performances of the data-driven approach are worsened. Nevertheless, due to the large confidence intervals, the hybrid model appears to be a reasonable compromise between the analytical and the GP model.  

	\section{Conclusions}\label{sec:Conclusions}
In order to have accurate control, a precise robot kinematic model is needed. The design and the actuation, however, may lead to complex systems, with many nonlinearities. Surgical robots, for example, are usually very articulated (eventually continuum) and tendon-driven, which gives rise to many nonliearities in the kinematic model. \\

Analytical methods allow to describe the robot kinematic model based on some assumptions and simplifications. Being based on mathematical formulations, they are usually very generalizable. On the other hand, data-driven approaches allow to build much more complex models from the data acquired. Nevertheless, they are less generalizable and the results are highly affected by the acquired dataset. In this work, GPR has been used for the data-drievn modelling thanks to its ability to provide a confidence interval for the accuracy of the results.

Hybrid approaches combining both methods may leverage the advantages of both methods, thus providing better modelling. The method presented in this work allows to mix analytical and data-driven, giving more importance to one or another depending on the level of confidence of the GPR model: where the data-driven model is uncertain, the analytical model is preferred.
Different methods for building the data-driven models have been used (GP for error learning, GP with prior, GP without prior). Results show that learning the error between the acquired data and the provided analytical model provides better results. The poorest model is the one obtained from GPR without any prior knowledge.
When using the proposed hybridization method, results show that the final model is affected by the provided analytical model. Wrong analytical models may reult in poorer models. However, when a more precise analytical model is employed, the modelling errors are highly reduced.

As noticed from the experiments on the real robot, the proposed method is capable of giving more importance to the analytical model or to the data-driven one, depending on their accuracy with respect to the collected data. The effect of the hybridization, however, depends on the parameters of the weighting function.

Future work will focus on improving the hybridization with an improved way to set the parameters of the weighting function and, following, on employing the hybrid model to control the robot in trajectory tracking and surgical task automation.


	




	
	\bibliographystyle{IEEEtran}
	\bibliography{./myLibrary}
	
\end{document}